# RevealED: Uncovering Pro-Eating Disorder Content on Twitter Using Deep Learning

Jonathan Feldman

# Abstract

The Covid-19 pandemic induced a vast increase in adolescents diagnosed with eating disorders and hospitalized due to eating disorders. This immense growth stemmed partially from the stress of the pandemic but also from increased exposure to content that promotes eating disorders via social media, which, within the last decade, has become plagued by pro-eating disorder content. This study aimed to create a deep learning model capable of determining whether a given social media post promotes eating disorders based solely on image data. Tweets from hashtags that have been documented to promote eating disorders along with Tweets from unrelated hashtags were collected. After prepossessing, these images were labeled as either pro-eating disorder or not based on which Twitter hashtag they were scraped from. Several deep-learning models were trained on the scraped dataset and were evaluated based on their accuracy, F1 score, precision, and recall. Ultimately, the Vision Transformer model was determined to be the most accurate, attaining an F1 score of 0.877 and an accuracy of 86.7% on the test set. The model, which was applied to unlabeled Twitter image data scraped from "#selfie", uncovered seasonal fluctuations in the relative abundance of pro-eating disorder content, which reached its peak in the summertime. These fluctuations correspond not only to the seasons, but also to stressors, such as the Covid-19 pandemic. Moreover, the Twitter image data indicated that the relative amount of pro-eating disorder content has been steadily rising over the last five years and is likely to continue increasing in the future.

# Introduction

**Background**

The Covid-19 pandemic brought about a significant increase in the number of adolescents diagnosed with eating disorders and in the number of adolescents hospitalized due to eating disorders (Agostino et al., 2021; Matthews et al., 2021; Otto et al., 2021). This growth stemmed from the fact that the pandemic expanded the prevalence of factors that induce eating disorders, such as seclusion and stress, and decreased the availability of factors that would inhibit the development of eating disorders (Culbert et al., 2015). Moreover, the pandemic caused a near-universal growth in social media use (Parsons et al., 2021; Rosen et al., 2022), and since social media serves as an environment that not only fosters pro-eating disorder (Pro-ED) sentiment, but also spreads it worldwide (Branley & Covey, 2017; Sukunesan et al., 2021), Pro-ED content was able to reach more people than ever before. In turn, a greater number of people were exposed to these dangerous ideas and began to post similar content (Mento et al., 2021), growing the Pro-ED movement substantially.

Recently, due to the pandemic-induced rise in eating disorder cases, numerous studies were conducted analyzing a vast array of possible methodologies to mitigate the spread of Pro-ED content on social media. Among these studies, several have examined the possibility of applying artificial intelligence-based techniques to mitigate the spread of content that promotes eating disorders on social media (Fardouly et al., 2022; Ren et al., 2022). Many have already applied models such as the Bidirectional Encoder Representations from Transformers (BERT) model and its derivatives to determine whether text-based social media posts are promoting eating disorders (Benítez-Andrades et al., 2021, 2022; Oksanen et al., 2015). Despite the numerous studies of eating disorder-related content on social media and the vast improvements in image-based artificial intelligence algorithms, none have yet employed computer vision techniques to detect whether an image in a social media post promotes eating disorders. Images constitute a substantial portion of posts on social networking sites, thus making image analysis crucial to understanding the role of eating disorders online. In addition, images, specifically those with the intent to promote eating disorders, are capable of inducing body dissatisfaction and emotional distress in viewers, which can be detrimental to those suffering from an eating disorder (Spettigue & Henderson, 2004). Hence, image-based artificial intelligence techniques must be utilized to attenuate the influence of the pro-eating disorder movement on social media.

**Objectives**

This study had two objectives. The first objective was to create an accurate image classifier that could detect content that promotes eating disorders on social media solely from image data. The second objective was to use the trained classifier to analyze the content of Tweets over time to understand to what extent the ubiquity of pro-eating disorder content has grown on Twitter over the last five years and to uncover any underlying patterns within the Pro-ED movement on Twitter.

**Literature Review**

Supervised machine learning (ML) and deep learning (DL) methods, which can be used to predict an outcome based on input data, have flourished thanks to the availability of data on social media. There are many supervised learning techniques, such as regression and classification (Nasteski, 2017). All of them require a labeled dataset which is used to train, and, possibly, test the model. Supervised learning models can then be used to classify large swathes of data, saving time, and generating predictions of greater accuracy than humans could (Goh et al., 2020).

A previous study (Benítez-Andrades et al., 2022) created a dataset of over two thousand classified Tweets which were subsequently used to train and test several ML and DL models, all of which seemed to be highly accurate. Another study researched the potential benefits and limitations of artificial intelligence for the prevention and diagnosis of eating disorders, concluding that such models are highly effective (Fardouly et al., 2022). Several other studies have also used social media data to analyze the effect of eating disorder-related content from a social perspective instead of a quantitative one (Gordon, 2000; Khosravi, 2020; Parsons et al., 2021).

In short, previous studies have been successful in detecting posts that promote eating disorders (Pro-ED) on social media using natural language processing (NLP) techniques, showing that DL and ML techniques can accurately recognize Pro-ED text (Benítez-Andrades et al., 2021, 2022). This study differs from its predecessors because, rather than analyzing text posted on Pro-ED platforms, it analyzes images. On social media, images play a vital role in communication and the diffusion of ideas, therefore, image analysis offers crucial insights into the dispersal of Pro-ED content online.

# Methods

**Data Collection**

Publicly available Tweets from several Twitter hashtags were scraped, downloading not only the content of the tweet, but also information about the post, such as the tweet id, the number of likes, retweets, replies, and hyperlinks to any images associated with the post. This study uses Twitter to obtain the dataset because, unlike other social media platforms such as Instagram, Facebook, and Tumblr, it does not currently have a policy for censoring hashtags or posts that promote eating disorders. Twitter is, therefore, an ideal space for Pro-ED content to disseminate and is host to a massive Pro-ED community (Arseniev-Koehler et al., 2016; Bert et al., 2016; Sukunesan et al., 2021). Such an unregulated platform provides space for a wholistic analysis of unrestricted Pro-ED content, which is ideal for this study.

Tweets from hashtags that previous studies have shown to disseminate Pro-ED content were scraped in their entirety. These hashtags consisted of "#proana", "#thinspo", "#thinspiration", "#fitspiration", and "#fitspo". These specific hashtags were chosen because previous research has revealed them to spread Pro-ED content (Dignard & Jarry, 2021; Sukunesan et al., 2021). Images from these hashtags made up the "Pro-ED" section of the scraped dataset. Similarly, Tweets from hashtags not associated with eating disorders were also scraped. These Tweets encompass many diverse images including those of people, animals, and landscapes. Such hashtags included "#ootd", "#fakecandid", "#animals", "#pets", "#travel", and "#photography". These hashtags have not been shown to spread Pro-ED content and are not associated with the hashtags from the Pro-ED portion of the dataset. These are valid candidates for the Not-Pro-ED section of the dataset because they contain a wide variety of image subjects, which trains the model on many different image types, subsequently decreasing the chances that a model will misclassify an image.

Upon scraping the tweet metadata, the Twitter image links were extracted, and all images were downloaded and marked as Pro-ED or Not Pro-ED based on the hashtags from which they were scraped, following the categorization described earlier. Such an approach, while potentially open to misclassifications, is an efficient way to create large image datasets on which to train the classifier, which is of particular importance when analyzing the manifold content of social media.

**Data Cleaning**

In total, 18,822 images were scraped from Twitter. Since most of the images were associated with more than one of the scraped hashtags all images with identical image URLs were treated as duplicates and deleted from the dataset. To ensure all duplicates were removed, a dHash algorithm (Yang, 21) was used to remove all images that had a similarity of 90% or more (Chamoso, 2017). Following the removal of duplicate images, the combined dataset decreased to 13,448 images.

**Data Labeling**

The dataset was subsequently split into two distinct categories based on their associated hashtag. Images from hashtags that are documented to proliferate Pro-ED content, such as "#proana", were labeled as "Pro-ED", while images from Not Pro-ED hashtags were labeled as "Not Pro-ED". Images classified as "Pro-ED" were then relabeled as 0, while images classified as "Not Pro-ED" were relabeled as 1—a necessary step, since classification algorithms can only process numerical values.

# Classification Methods

**General**

  Two computer vision frameworks were chosen for this classification task, the Residual Neural Network (ResNet) model and the Vision Transformer (ViT) model. The ResNet algorithm was chosen because it has been the default Convolutional Neural Network (CNN) framework since it won the ImageNet Large Scale Visual Recognition Challenge in 2015. The ResNet algorithm also serves as a baseline computer vision model to which other image recognition algorithms can be compared since it has been heavily studied and documented (He et al., 2015; Li et al., 2017; Targ et al., 2016; Wu et al., 2019). The ViT algorithm was chosen because it utilizes a transformer-based framework (Dosovitskiy et al., 2021), unlike the standard CNN, which reflects the current state-of-the-art in image recognition (Yu et al., 2022). Both frameworks were initially trained on the ImageNet-1k dataset (Deng et al., n.d.), which consists of over 1.2 million images. To classify Tweets, each model was individually fine-tuned on the training dataset. Fine-tuning involves freezing all layers of the neural network except the final fully connected layer and then changing only the weights of the non-frozen final layer to create a classifier suitable for the necessary task. Fine-tuning allows the classifier to utilize pretrained weights from a large dataset such as ImageNet-1k, which aids in recognizing features in images, leading to greater classification accuracy (Tan et al., 2018; Zhou et al., 2017). Both models were trained on the same combined dataset, of which 80% of the images were used for training and validation, while 20% of the images were used for testing. Each model was fine-tuned for 20 epochs. The weights from the epoch that achieved the highest validation accuracy were used in the final model.

**ResNet**

The ResNet model (Rao et al., 2021) follows the classic CNN structure which is characterized by adjoining groups of convolutional layers attached together by a pooling layer. Most CNNs end with a group of fully connected linear layers that output the final classification (He et al., 2015). The basic structure of ResNet was not a new development. It was first seen in the LeNet-5 image classifier which was originally published in 1998. What differentiates ResNet from its predecessors is its approach to mitigating the vanishing gradient problem—a common issue in deep neural networks since the gradient continually decreases as it propagates backward through the model until it all but vanishes. The ResNet architecture alleviates the vanishing gradient problem by providing gradients a means by which they can skip over layers. The gradient then benefits by maintaining its value for longer. For this task, a variant of ResNet, ResNet-152, was used. ResNet-152 is already pretrained and is available through the PyTorch model library. This model differs from other ResNet models due to its increased number of convolutional/pooling layer groups. As a result of the depth of ResNet-152, it outperforms most ResNet variants (Han et al., 2018), while maintaining the integrity of its gradient due to the skip connections.

**Transformer**

Transformers—an integral part of Vision Transformer models—are a deep learning model that uses a self-attention mechanism to impute significance to any given section of input data (Vaswani et al., 2017). Self-attention allows transformers to draw information from pieces of input data that came prior, providing crucial knowledge of the relationship between different inputs within the sequence. Transformers analyze sequential data but process the entire input all at once. During processing the self-attention mechanism establishes contextual links between the different parts of the input data, such as words in an input statement. However, due to this self-attention processing, transformers can process multiple input pieces at a time, allowing for parallelization and reducing time spent on computation (Rao et al., 2021).

**Vision Transformer**

The Vision Transformer (ViT) model differs greatly from the past CNN structures because it employs transformers rather than a sequence of convolutional and pooling layers. ViT, inspired by transformer-based NLP models (Dosovitskiy et al., 2021), leverages a transformer-based architecture for computer vision tasks, achieving a far better performance than regular CNNs. The self-attention mechanism within transformers allows ViT to avoid the common inductive biases of standard CNNs by simultaneously analyzing the entire input image. In essence, transformer-based computer vision models can extract not only visual features from an image but also interdependencies between different parts of an image, by analyzing the image wholesomely (Islam, 2022; Jeeveswaran et al., 2022).

Unlike ResNet, ViT does not follow the standard CNN structure. Alternatively, the model begins by splitting up an input image into flattened patches and transforming the patches into sequential linear embeddings. These embeddings are then inputted into the transformer encoder which alters the data based on learned interdependencies. Afterward, the encoded data is fed through a multilayer perceptron (MLP) which generates a classification.

**Procedure**

The image dataset was organized, loaded, and transformed through PyTorch's built-in imagefolder class. The ResNet-152 model was instantiated using PyTorch's torchvision.models' package. The ViT Model was downloaded from the Google AI's Hugging Face model repository.

The ViT model was finetuned through the Hugging Face API in addition to PyTorch's computer vision library: Torchvision.

# Results

**Performance**

The fine-tuned ViT model was found to be more accurate on the test set than the ResNet model. ViT achieved approximately 86.7% accuracy and received an F1 score of 0.877. The ResNet model achieved 83.2% accuracy and received an F1 score of 0.837. The difference in accuracy and F1 score between the two models is expected since research has indicated that ViT models consistently outperform ResNet models (Chen et al., 2022). Moreover, ViT also routinely attained faster training and inference speeds than ResNet-152, as can be seen in the Error vs. Epoch graphs in Table 1 which graphs the models' accuracy, relative to the correct prediction, over the period of training time, otherwise known as epochs. Such graphs illustrate the gradual improvement of a model's accuracy as it "learns" from the data.

Both the ResNet-152 and the ViT model have a significantly higher recall than precision, indicating that both models are more likely to generate false positive classifications than false negative classifications.

**Analysis**

After concluding that the ViT model outperforms the ResNet model, a method to test the efficacy of the model on large social media datasets was devised. To examine the ViT model's capabilities, twitter images from "#selfie" were scraped. It was determined that stratified random sampling would be the most effective sampling method because precise conclusions can be made about certain subgroups, in this case, months, from stratified sampling without the need for large datasets (Howell et al., 2020; Shadish et al., 2002). For a given month, three random non-consecutive days were chosen, and for each of the three chosen days, all images posted on the chosen hashtags were scraped.

**Table 1**

*Error vs. Epoch Graphs*

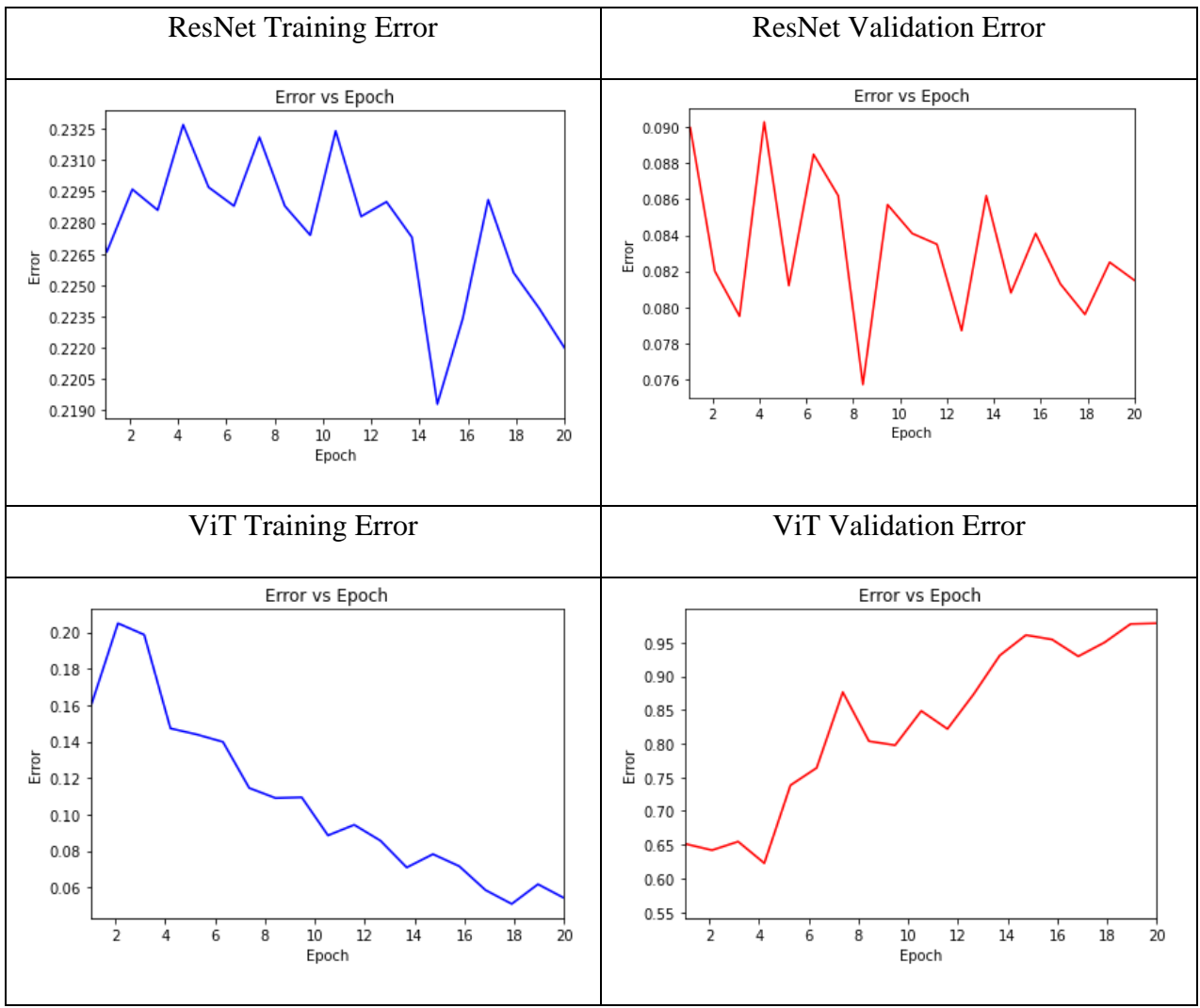

*Note.* The blue colored graphs measure the training error of each model, while the red colored graphs measure the validation error of each model.

The process was repeated for each month between January 2017 and June 2022. The hashtag, "#selfie", was chosen because most of the images posted on it are focused on human beings and their appearance, mimicking the content posted on Pro-ED hashtags. In total, 56,445 images were scraped for analysis from "#selfie". All the scraped images were then aggregated by month and then fed into the ViT model, which generated a classification. For each month, the classifications for the individual images were used to calculate the percentage of Pro-ED images posted within the time frame. These percentages were then arranged by date and graphed.

**Figure 1**

*Linear Regression Data Fit*

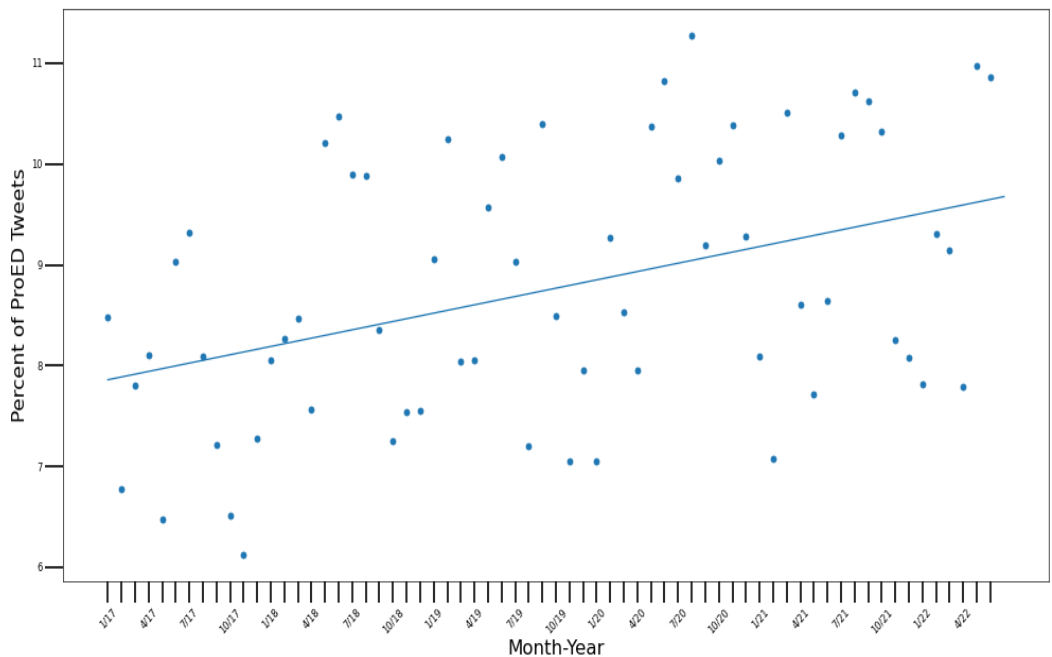

**Figure 2**

*Polynomial Regression Data Fit*

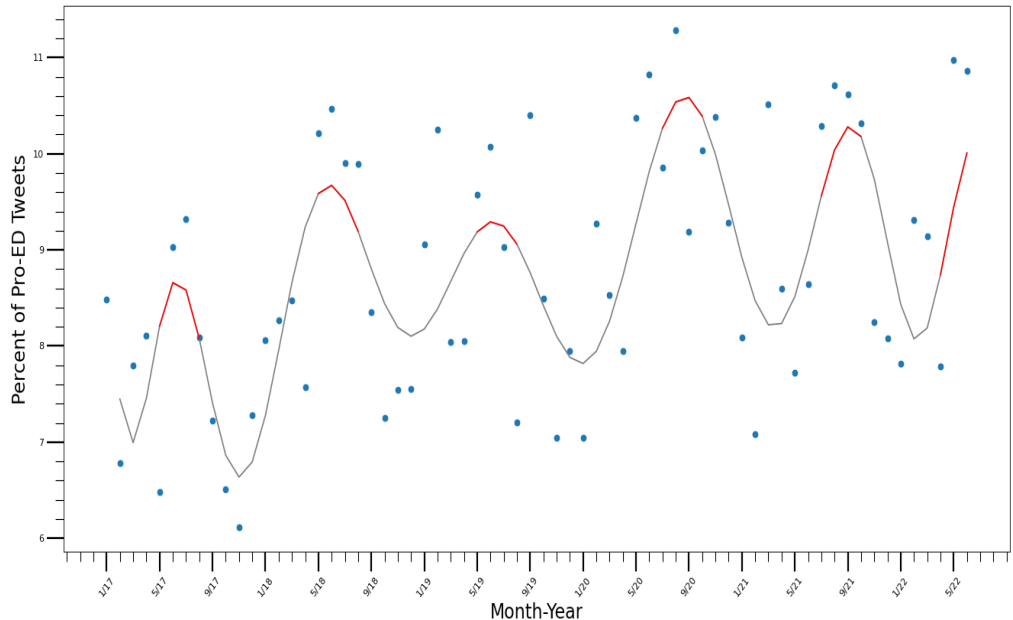

From this graphed data, a trend line was generated, shown in Figure 1. The trend line indicates that there is a positive correlation between the progression of time and the relative amount of Pro-ED images on "#selfie". However, the r-squared value of the trend line is 0.166. Such a small r-squared value reveals that the trend line does account for a vast majority of the variation in the data. To remedy this, a polynomial regression was fit to the data, as shown in Figure 2. The generated regression had an r-squared value of 0.555, indicating that the polynomial graph illustrates most of the data's variations. Moreover, the RMSE of the linear regression is 1.31, while the RMSE of the polynomial regression is 0.871, once again indicating that a polynomial regression provides a superior fit. Both the linear regression and the polynomial regression have holistic P values of less than 0.001 (P < .001), which indicates that both regression models are statistically significant.

**Figure 3**

*Aggregate Mean of Pro-ED Content Per Month From 2017-2022*

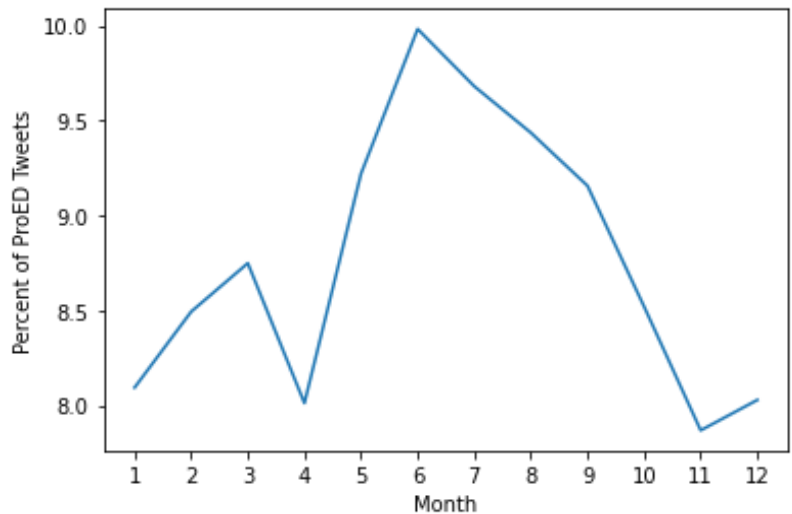

*Note.* This figure is the graph of the mean percent of Pro-ED content on "#selfie" during each month based on every year of collected data.

The data extracted from "#selfie" suggests that there is an annual increase in the relative number of Pro-ED images posted during late spring and early summer (April to June). Furthermore, the data indicates that the relative number of Pro-ED images posted on "#selfie" reaches its highest value in the late summer or early autumn (July to September) every year, as is shown in Figure 3. Moreover, the data implies that the relative amount of Pro-ED posts on "#selfie" has been gradually increasing every year since 2017. The mean and standard deviation of the relative number of Pro-ED posts have also been growing since 2017. The growth in mean and standard deviation stipulates that over the last five years, the percentage, relative to the whole, of Pro-ED content has been increasing, and that the difference between the relative total of Pro-ED content in the summer and during other times of year has been growing as well. Based on the polynomial and the linear regressions, it is likely that this growth will continue into the next year and beyond.

# Discussion

**Principal Findings**

This study explored the efficacy of training a deep learning model to classify Pro-ED content on the social media platform Twitter. Two computer vision models, ViT and ResNet, were trained using scraped Tweets and then assessed to determine their accuracy on a hold-out test set. The results of this analysis indicated that both models were highly effective at recognizing Pro-ED content. Though both models do exceedingly well at recognizing Pro-ED content, they are both hindered by the scope of the training data which causes them to generate false positive classifications. One likely explanation for the higher number of false positive classifications in comparison to false negative classifications is the presence of Pro-ED content within the Not Pro-ED set and test set. Such data overlap would cause a greater number of false positives and is indeed possible due to the unedited nature of the dataset.

Next, after training, the ViT model was deployed to analyze a vast number of images from Twitter's "#selfie". The results of the ViT model's analysis affirmed long-standing theories which relate worsening eating disorder symptoms to specific seasons of the year. Previous research has indicated that those who suffer from certain eating disorders—largely bulimia—see an annual increase in eating disorder severity during certain seasons (Fornari et al., 1994; Ghadirian et al., 1999; Yamatsuji et al., 2003). These seasonal changes in symptom intensity are often attributed to Seasonal Affective Disorder (SAD), which is a disorder characterized by seasonal patterns of major depression followed by a period of remission (Hardin et al., 1991; Nasteski, 2017). Though it was previously believed that SAD was reserved solely for winter, recent research has uncovered that for those suffering from body-image concerns, it may also occur during the summer (Akram et al., 2019; Ghadirian et al., 1999; Liang et al., 2018).

The analysis of "#selfie" illustrated similar seasonal fluctuations in Pro-ED sentiment—measured by the relative amount of Pro-ED content—to those described in studies that analyzed the seasonality of eating disorder symptoms (Ghadirian et al., 1999; Hardin et al., 1991; Melrose, 2015). In this instance, during late spring and early summer, the relative number of Pro-ED images on "#selfie" increases, reaching its peak during July or August. Afterward, the relative abundance of Pro-ED images falls, reaching its lowest point in autumn or early winter. This wave-like pattern consisting of an annual rise and fall in the relative number of Pro-ED content on "#selfie" is analogous to the seasonal fluctuations of SAD and is supported by research which shows that eating disorder symptoms, and, in turn, Pro-ED sentiment increases during the summer (Sloan, 2002). Moreover, a previous study determined that people's mental well-being deteriorates during the summer and that social media reflects this deterioration through an increase in the amount of depressive language in posts (Burke et al., 2018). Since eating disorders and depression are closely linked, worsening symptoms of depression during the summer could also bring about an increase in eating disorder severity (Garcia et al., 2020; Mischoulon et al., 2011; Presnell et al., 2009;

Sander et al., 2021), and therefore, dramatically enlarge the relative abundance of Pro-ED content, as is seen on "#selfie".

Recent research has shown that the Covid-19 pandemic worsened the condition of eating disorder patients and led to an increase in new eating disorder cases (Khosravi, 2020; Rodgers et al., 2020; Taquet et al., 2022). The foregoing data from "#selfie" depicts a similar story. In January 2020, which marked the beginning of the global pandemic, there was a sudden increase in the relative amount of Pro-ED content. However, this sudden increase continued, and the number of Pro-ED images kept on rising all through spring and summer. By August 2020, the relative number of Pro-ED posts on "#selfie" had outgrown all years prior. To this day, the year 2020 boasts the highest mean Pro-ED content of all time on "#selfie". Despite the end of the global pandemic, the relative number of Pro-ED images on "#selfie" has not declined to its pre-pandemic levels. Moreover, the data does not indicate any future decline.

The ViT model's classifications of over four years' worth of images extracted from "#selfie" illustrated seasonal trends in content promoting eating disorders and an overall increase in said content over the last four years. These findings are analogous to the trends uncovered in previous studies on both the seasonality of eating disorders and the effect of social media on the spread of Pro-ED content. The congruency between the results of the ViT model and non-artificial intelligence-based studies of the same phenomenon indicates that the model can recognize and understand the nuances of Pro-ED images, differentiating it from other image types, even when analyzing swathes of data. Moreover, the ViT model reaches accuracies comparable to NLP-based models developed in previous studies, even though the ViT model analyzes an entirely different informative medium (Benítez-Andrades et al., 2021; Oksanen et al., 2015). Most importantly, like other artificial intelligence models, the model developed in this study can analyze images at speeds and accuracies that humans are unable to replicate, offering a quick and efficient method for recognizing Pro-ED content on social media.

There does not exist a perfect solution for halting the spread of Pro-ED content on social media. However, computer vision models, such as ViT, and NLP models, such as BERT, offer a possible countermeasure (Astorino et al., 2020; Benítez-Andrades et al., 2022). These models can be employed by social media companies to detect and remove content that the models classify as Pro-ED, or they can be used by external organizations to monitor the prevalence of Pro-ED content on social media. In any case, these models offer an efficient and accurate method of detecting Pro-ED content on social media and may be used to mitigate the spread of the Pro-ED movement, thereby restricting the harmful effects of such posts (Feldhege et al., 2021).

**Limitations**

There are several limitations in the research conducted. First, the dataset is comprised of images from a single social media platform. Second, the images in the datasets were labeled based on the hashtag from which they were downloaded and, therefore, there are potentially images that have an incorrect classification. Third, the Pro-ED hashtags that were scraped to build the dataset are comprised only of those Pro-ED hashtags that have been documented to support eating disorders. Images from many other Pro-ED hashtags have been omitted from the dataset.

# Conclusion

Deep Learning computer vision models were trained to classify images extracted from Tweets as either promoting eating disorders or not. Both models achieved accuracies of greater than 83%. The best-performing model was the Vision Transformer model, a model which employs a transformer-based architecture to generate a classification based on an input image.

The Vision Transformer model was used to analyze Tweets that were posted on "#selfie" over the last five years. The analysis uncovered seasonal patterns of increased pro-eating disorder presence on social media during the summer season. Thus, providing much-needed empirical evidence to support earlier theories about seasonal fluctuations in eating disorder symptoms. Moreover, the Vision Transformer model also revealed the profound effect the Covid-19 pandemic has had on the prevalence of Pro-ED content on social media and revealed how the presence of stressors influences the abundance of Pro-ED content.

Future research will focus on increasing the current dataset to elevate model generalizability and erase any incidental bias, pairing image processing and natural language processing techniques to create a model capable of analyzing any social media post, and utilizing the trained model in the real world, potentially as a bot, to uncover and report posts and platforms that promote eating disorders.